\documentclass[runningheads]{llncs}
\usepackage{hyperref}
\usepackage[T1]{fontenc}
\usepackage{cite}

\usepackage{graphicx}
\usepackage{array} 
\usepackage{textcomp}
\usepackage{multirow}

\newcommand{\smallplus}{\raisebox{0.2ex}{\scalebox{0.8}[0.8]{$+$}}}
\newcommand{\smalltilde}{\raisebox{0.2ex}{\scalebox{0.8}[0.8]{$\sim$}}}

\begin{document}
\title{Predicting and Interpolating Spatiotemporal Environmental Data: A Case Study of Groundwater Storage in Bangladesh}

\titlerunning{Predicting and Interpolating Spatiotemporal Environmental Data}

\author{Anna Pazola\inst{1,2} \and
Mohammad Shamsudduha\inst{3} \and
Richard G. Taylor\inst{2}
\and Allan Tucker\inst{1}}
\authorrunning{A. Pazola et al.}
%
\institute{Department of Computer Science, Brunel University of London, UK \and
Department of Geography, UCL, London, UK \and
Department of Risk and Disaster Reduction, UCL, London, UK \\
\email{anna.pazola@brunel.ac.uk} \\
}
\maketitle              
\begin{abstract}
Geospatial observational datasets are often limited to point measurements, making temporal prediction and spatial interpolation essential for constructing continuous fields. This study evaluates two deep learning strategies for addressing this challenge: (1) a grid-to-grid approach, where gridded predictors are used to model rasterised targets (aggregation before modelling), and (2) a grid-to-point approach, where gridded predictors model point targets, followed by kriging interpolation to fill the domain (aggregation after modelling). Using groundwater storage data from Bangladesh as a case study, we compare the efficacy of these approaches. Our findings indicate that spatial interpolation is substantially more difficult than temporal prediction. In particular, nearest neighbours are not always the most similar, and uncertainties in geology strongly influence point temporal behaviour. These insights motivate future work on advanced interpolation methods informed by clustering locations based on time series dynamics. Demonstrated on groundwater storage, the conclusions are applicable to other environmental variables governed by indirectly observable factors. Code is available at \url{https://github.com/pazolka/interpolation-prediction-gwsa}.

\end{abstract}
\section{Motivation}
Spatial interpolation and temporal prediction are two modelling objectives that go hand-in-hand in environmental modelling and monitoring \cite{cressie2011statistics}. Because continuous data collection is costly and often infeasible, many datasets are only available at point locations, hindering various downstream tasks that rely on them, including planning, risk assessment and decision making.

This study explores how these two tasks can be combined within data-driven models to pursue the "holy grail" of spatiotemporal modelling: estimating values at unobserved locations and future times. What are the existing approaches, especially deep learning-based solutions, for addressing this problem? Using an extensive spatiotemporal dataset on groundwater storage changes in Bangladesh, we identify key challenges and report performance of selected models, offering insights relevant for other environmental variables with complex and poorly understood governing physical processes. 

As machine learning (ML) research and AI industry develops, primarily driven by large language models, the environmental cost of AI is widely recognised and discussed, e.g. \cite{bolon-canedo_review_2024}. In this work, we focus on lightweight models that contribute to the discourse on simpler, and more interpretable deep learning approaches, closely integrated with domain expertise, as an alternative to to general-purpose, transformer-based models that are computationally expensive and environmentally costly \cite{rudin_amazing_2024}. 

\section{Background}

Estimation of continuous fields from observational datasets involves spatial interpolation and temporal gap filling.
In geostatistics, Kriging is the best linear unbiased predictor (BLUP) and is one of the most commonly used interpolation techniques \cite{cressie2006spatial}. It takes into account not only distance but also correlation between neighbouring points. However, with its strong assumptions on Gaussianity and stationarity, and very high computational cost due to a creation of large distance matrix (\(O(n^{3})\) for time), it is often unsuitable for contemporary, large datasets \cite{cressie2006spatial}. Common approaches for temporal gap filling vary depending on the characteristics of time series, with autoregressive integrated moving average (ARIMA) and seasonal-trend decomposition-based methods, as well as ML models (e.g. random forest \cite{arriagada_automatic_2021}) being popular choices. Merging these tasks might leverage the quality of estimations, by taking into consideration spatial and temporal dependencies \cite{cressie2006spatial}.

A variety of models are well-suited for both spatiotemporal interpolation and short-term extrapolation tasks. These can be broadly categorized into groups: 
\begin{itemize}
    \item \textit{Statistical methods}: This includes space-time (3D) kriging \cite{li_temperature_2020}, and its ML counterpart, spatiotemporal DeepKriging \cite{nag_spatio-temporal_2023}.
    \item \textit{Deterministic generative models}: Examples include 2D/3D U-Net architectures \cite{kishikawa_conditional_2025} and CNN-LSTM stacks. These models typically require prior rasterisation of the target data, or a subsequent interpolation step, for generating gridded outputs.
    \item \textit{Probabilistic models}: These encompass spatial Gaussian processes (GP) - mathematically equivalent to kriging and similarly computationally intensive \cite{zhang_efficient_2022} - as well as neural processes \cite{garnelo_neural_2018}.
\end{itemize}

In this work, we analyse how different combinations of prediction and interpolation tasks in the selected methods (U-net variants, CNN-LSTM models, spatiotemporal DeepKriging) affect the estimation of a spatiotemporal variable at unobserved locations.

\section{Methods}
\label{methods}
The method selection process prioritised computational efficiency and ability to handle gridded predictors and point targets with minimal pre- and postprocessing, aiming to generate spatially and temporally complete datasets for a specified area and time resolution.

Two strategies were used for handling the format of observed targets: rasterising point observations before modelling (grid-to-grid approach), or after modelling via spatial interpolation (grid-to-point model with subsequent interpolation step). In each case, we explored 2D (purely spatial) and 3D (spatio-temporal) versions of the models.

\subsection{Grid-to-grid approach: Unet}
Point observations can be rasterised by aggregating all points that fall within the same grid cell. Although the resulting target grid contains gaps, the modelling task can be framed as a 2D/3D image inpainting problem, using spatially complete gridded predictors as input. We employed a variant of U-net, which has been successfully applied for similar problems \cite{meuer_infilling_2025}. When built with 3D layers (i.e. 3D convolutions, max-pooling and upsampling), the network can process temporal sequences, enabling it to capture time-dependent relationships (time lags) that are characteristic of environmental variables.

The number of encoder-decoder blocks, consisting of 2D or 3D Convolutions, MaxPooling and UpSampling steps for purely spatial U-net and space-time U-net respectively, is flexible and mostly independent from the input shape. However, the network depth should reflect the complexity and size of the inputs, i.e. shallower U-nets might be more appropriate for small inputs, reducing the number of parameters and avoiding overfitting.

To handle missing spatial data, the model uses masked mean squared error (MSE) losses that compute errors only over valid pixels. In the 2D U-net, masked loss averages per-pixel MSE over valid regions, ignoring masked areas. In the 3D U-net, masked loss adds penalties for low-frequency (trend) and spatial-mean differences between predictions and targets:
\begin{equation}
    Loss=w_{main}L_{MSE}+w_{trend}L_{lowpass}+w_{mean}L_{mean}
\end{equation}
where $L_{MSE}$: standard masked per-pixel MSE, $L_{lowpass}$: masked MSE of low-frequency (trend) components obtained via normalized average pooling, $L_{mean}
$: masked MSE between spatial means of true and predicted fields. This encourages the U-net to match both local details and large-scale trends, improving stability and physical consistency in the 3D (space–time) setting.

\subsection{Grid-to-point approach: CNN/CNN-LSTM stack}
As opposed to rasterising point observations before modelling, we explored the effects of splitting the interpolation and prediction tasks, and build a predictive grid-to-point model, followed by a separate interpolation step. The model consists of two (time-distributed) convolutional layers and max-pooling pairs, followed by a dense layer in the spatial version, or two LSTM layers and a dense layer in the spatio-temporal version. The output size corresponds with the number of all point observations, therefore it predicts the target at all observed locations simultaneously.

\subsubsection{Kriging}
To extend point predictions to the whole study area, ordinary kriging was applied for the subsequent interpolation step, for each time step independently. The choice of an appropriate variogram model, fitted empirically to the data, is crucial for accurately representing spatial correlation and ensuring the validity of the interpolation results.

\subsection{Spatiotemporal DeepKriging}
We decided to compare our models to spatiotemporal DeepKriging, that conceptually echoes 3D kriging by treating time as an additional spatial dimension within a deep learning framework \cite{nag_spatio-temporal_2023}. It separates interpolation (stage 1) from temporal prediction (stage 2), which opposes our CNN/CNN-LSTM setup. Stage 1 performs interpolation at observation points using RBF-based embeddings of space and time coordinates, combined with scalar predictors. A feed-forward network trained with a quantile loss produces multiple predictive quantiles, enabling uncertainty estimation. Stage 2 then carries out grid-based forecasting with a ConvLSTM that ingests recent spatiotemporal sequences and full-grid predictors to predict future quantile fields, using a masked quantile loss to ignore invalid regions. Both stages rely on separate models that are trained independently.

\section{Case study: groundwater storage changes in Bangladesh}

Estimating changes in groundwater storage ($\Delta$GWS) on a continuous grid is critical for informing sustainable water management from regional to local scales. Typically derived from a combination of land surface model outputs and satellite gravimetry, $\Delta$GWS exhibit considerable uncertainty \cite{gleeson_gmd_2021, arifin_plausibility_2025}. Many areas across the world, such as Bangladesh, rely heavily on groundwater for drinking water supplies and national food security through dry-season irrigation \cite{shamsudduha_impacts_2018}. However, as documented in Bangladesh, excessive groundwater abstraction has led to localised depletion, most importantly in Dhaka and western regions \cite{janardhanan_groundwater_2023}. Taken together with the existing water-quality challenges (salinity, arsenic), these pressures make robust groundwater risk management indispensable, especially as global warming puts considerable pressure on water resources \cite{shamsudduha_impacts_2018}.

Bangladesh has an extensive network of point observations of groundwater levels that can be used as a basis for data-driven modelling on the country scale. A set of gridded satellite-based predictors (precipitation, actual evapotranspiration, NDVI, elevation, terrestrial water storage changes) was selected based on the forcing variables used commonly in the literature, e.g. \cite{faruki_fahim_modeling_2024}.

\begin{figure*}[]
  \centering
  \includegraphics[width=0.95\textwidth]{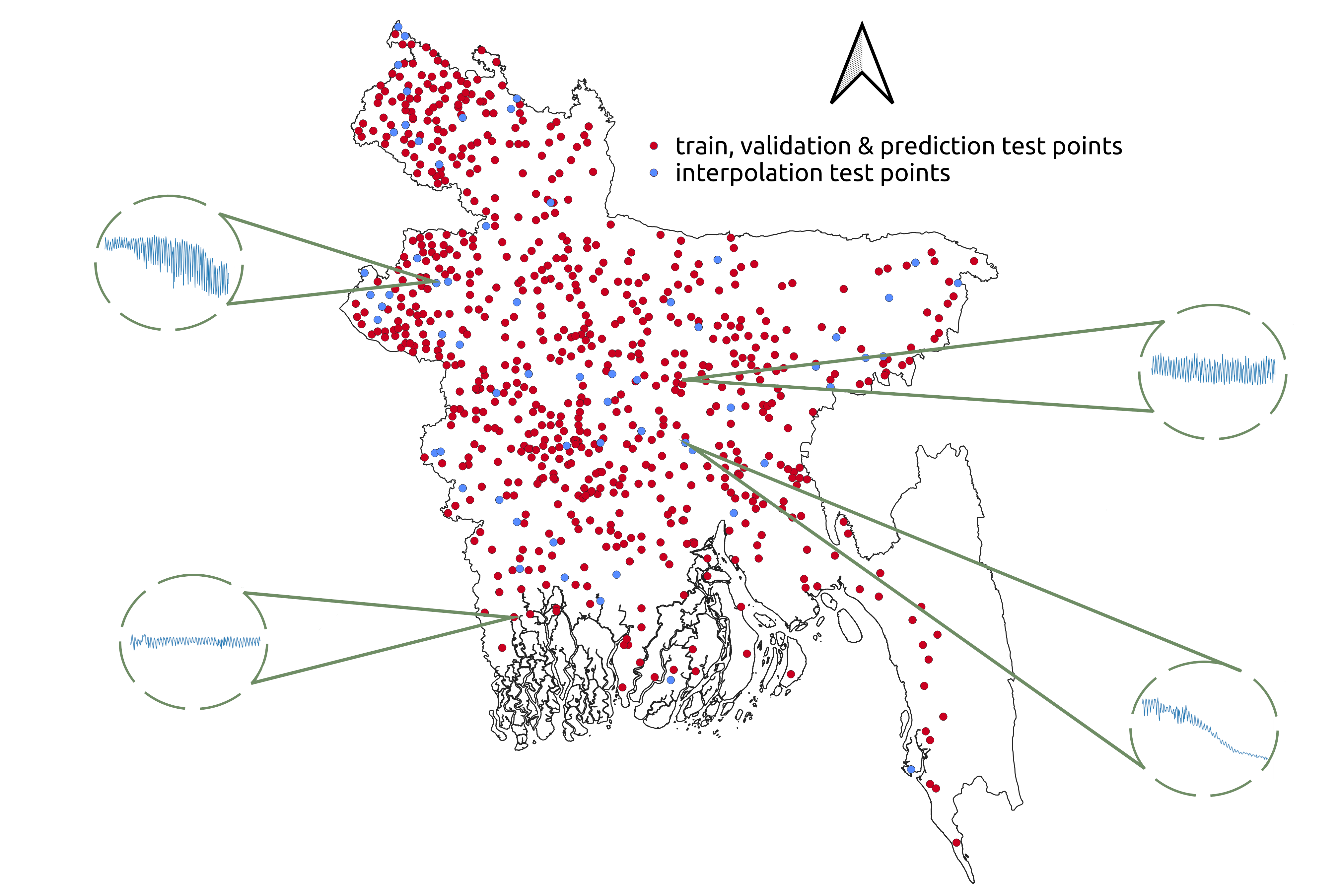}
  \caption{Distribution of train, validation \& prediction test points, and interpolation test (holdout) points. Groundwater dynamics in Bangladesh are characterised by a high spatial heterogeneity.}
  \label{fig:dist}
\end{figure*}

\subsection{Input and target datasets}

All geospatial datasets use the WGS84 geographic coordinates system. For kriging purposes, the coordinates were locally projected to WGS 84/UTM zone 46N.

\subsubsection{Gridded predictors} 
The gridded input data consists of the remotely sensed variables: MODIS’ NDVI as a proxy for agricultural output and irrigation, TERRACLIMATE precipitation and actual evapotranspiration as key meteorological variables controlling the available amount of water, MERIT DEM elevation, and GRACE $\Delta$TWS \cite{Save2020CSRMascon}. Input grids were reprojected to the common grid (0.25$^\circ$ x 0.25$^\circ$), and the temporal resolution of the dataset was set to 1 month. The study period is dictated by the availability of GRACE data (04/2002 - 05/2017).

\subsubsection{Groundwater level observations}
The target dataset consists of point observations of groundwater levels (GWL) at 1023 locations in Bangladesh (Fig. \ref{fig:dist}). Original weekly monitored GWL
time-series dataset can be obtained from the Bangladesh Water Development Board\footnote{\url{http://www.hydrology.bwdb.gov.bd/index.php}} though an online data request and payment. Processed monthly time-series dataset used in this work was compiled by \cite{shamsudduha_bengal_2022}.

\subsubsection{GLDAS 2.2 model output}
NASA's Global Land Data Assimilation System (GLDAS) data products provide a wide range of simulated hydrological variables, commonly used for estimating $\Delta$GWS. In particular, GLDAS 2.2 Catchment Land Surface Model CLSM-F2.5 \cite{li_global_2019} simulates shallow groundwater storage explicitly. In this work, average monthly outputs at a native daily scale at 0.25$^\circ$ were evaluated against available in-situ observations, alongside the outputs of the data-driven models.

\subsection{Preprocessing}
The raw target dataset required careful curation, as it contained anomalous values linked to human error and unreported measuring well changes, i.e. replacement of the existing observational well with a different well in a close proximity that might have a different depth and tap into a different aquifer. Such well changes were evident in some GWL time-series as sudden changes in the magnitude of seasonality, range of reported values and trends, and identified subseries were treated independently.

As most of the employed methods were unable to handle temporal gaps, with the exception of DeepKriging, temporal gap filling based on random forest regression was performed on GWL time series using \textit{R} package \textit{missForest} \cite{missForest, arriagada_automatic_2021}. Subsequently, GWL values were converted to groundwater storage using an aquifer storage coefficient dataset compiled by \cite{shamsudduha_bengal_2022}, and normalised w.r.t. 2004-2009 for direct comparison with GRACE and GLDAS anomaly values. Thus, the target values are denoted as $\Delta$GWS and are expressed in meters. For grid-to-grid models, rasterisation was performed through averaging of point values to a 0.25$^\circ$ x 0.25$^\circ$ grid.

{
\renewcommand{\arraystretch}{1.2}   
\setlength{\extrarowheight}{1pt}     
\setlength{\tabcolsep}{1.75pt}
\begin{table}[h]
\centering
\caption{Performance on $\Delta$GWS dataset from Bangladesh, R$^2$ (mean and standard deviation) over 10-fold temporal CV. Train-val-test refer to the temporal splits. The interpolation holdout locations are independent from the train-val-test locations (Fig. \ref{fig:dist}) and span the whole study period. The division of interpolation set into train-val-test results aims to isolate the "interpolation\smallplus prediction" performance, where values are predicted in the future at unobserved locations. Grid-to-point (g2p) without additional interpolation is also reported. All models were tested with 2D (spatial only) and 3D (spatiotemporal) inputs.}
\label{tab:performance}
\small
\begin{tabular}{lcccccccccccc}
\hline
\multicolumn{1}{c}{\multirow{3}{*}{\Large$R^2$}}  & \multicolumn{6}{c}{\textbf{prediction}}                                                                                                                                                                      & \multicolumn{6}{c}{\textbf{interpolation}}                                                                                                                                                                   \\
\multicolumn{1}{c}{\textbf{}} & \multicolumn{2}{c}{\textbf{train}}                                 & \multicolumn{2}{c}{\textbf{val}}                                   & \multicolumn{2}{c}{\textbf{test}}                                  & \multicolumn{2}{c}{\textbf{train}}                                 & \multicolumn{2}{c}{\textbf{val}}                                   & \multicolumn{2}{c}{\textbf{test}}                                  \\
\multicolumn{1}{c}{\textbf{}} & \multicolumn{1}{c}{mean} & \multicolumn{1}{c}{std} & \multicolumn{1}{c}{mean} & \multicolumn{1}{c}{std} & \multicolumn{1}{c}{mean} & \multicolumn{1}{c}{std} & \multicolumn{1}{c}{mean} & \multicolumn{1}{c}{std} & \multicolumn{1}{c}{mean} & \multicolumn{1}{c}{std} & \multicolumn{1}{c}{mean} & \multicolumn{1}{c}{std} \\ \hline
DeepKrig          & 0.66                            & 0.02                            & 0.57                            & 0.01                            & 0.02                            & 0.28                             & 0.61                           & 0.03                              & 0.46                           & 0.16                            & -0.02                           & 0.31                            \\
GLDAS 2.2            & 0.08                            & -                            & -0.05                           & -                            & -0.11                           & -                             & 0.08                            & -                           & -0.8                           & -                           & -0.44                           & -                           \\
2D Unet                 & 0.60                            & 0.04                            & 0.44                            & 0.12                            & 0.48                            & 0.08                             & 0.58                           & 0.04                            & 0.38                           & 0.17                            & 0.45                           & 0.12                            \\
3D Unet           & 0.58                            & 0.07                            & 0.45                            & 0.01                            & 0.45                            & 0.06                             & 0.57                           & 0.07                            & 0.41                            & 0.14                            & 0.44                           & 0.10                            \\
2D g2p\smallplus Krig          & 0.65                            & 0.02                            & 0.50                            & 0.13                            & 0.49                            & 0.10                            & 0.61                          & 0.03                              & 0.40                           & 0.17                            & 0.43                           & 0.13                          \\
3D g2p\smallplus Krig   & 0.65                            & 0.02                            & 0.49                            & 0.12                            & 0.49                            & 0.07                             & \textbf{0.61}                            & \textbf{0.03}                              & \textbf{0.41}                          & \textbf{0.16}                            & \textbf{0.44}                           & \textbf{0.11}                            \\
2D g2p        & \textbf{0.87}                            & \textbf{0.02}                            & \textbf{0.69}                            & \textbf{0.14}                          & \textbf{0.65}                                                     & \textbf{0.09}                                & -                                & -                               & -                                & -                               & -         & -                       \\
3D g2p  & 0.85                            & 0.02                           & 0.67                            & 0.12                            & 0.64                            & 0.07                             & -                               & -                                & -                               & -                                & -                               & -                       \\ \hline        
\end{tabular}
\end{table}
}

\subsection{Experiment setup}
To test both prediction and interpolation, a holdout set was set aside at the start, consisting of 8\% of the available locations selected randomly. Next, 10-fold temporal cross validation (CV) was performed on the remaining data, splitting the time series into train, validation and test sets, with a progressively growing training set, and validation and test sets of fixed length of 8 time steps. Unlike standard CV, successive training sets are supersets of those that come before them. Target and predictors were min-max scaled before modelling. Performance of the models were reported using the coefficient of determination R$^2$ and mean squared error MSE.

\section{Results and discussion}
\begin{figure}[!t]
  \centering
  \includegraphics[width=0.95\textwidth]{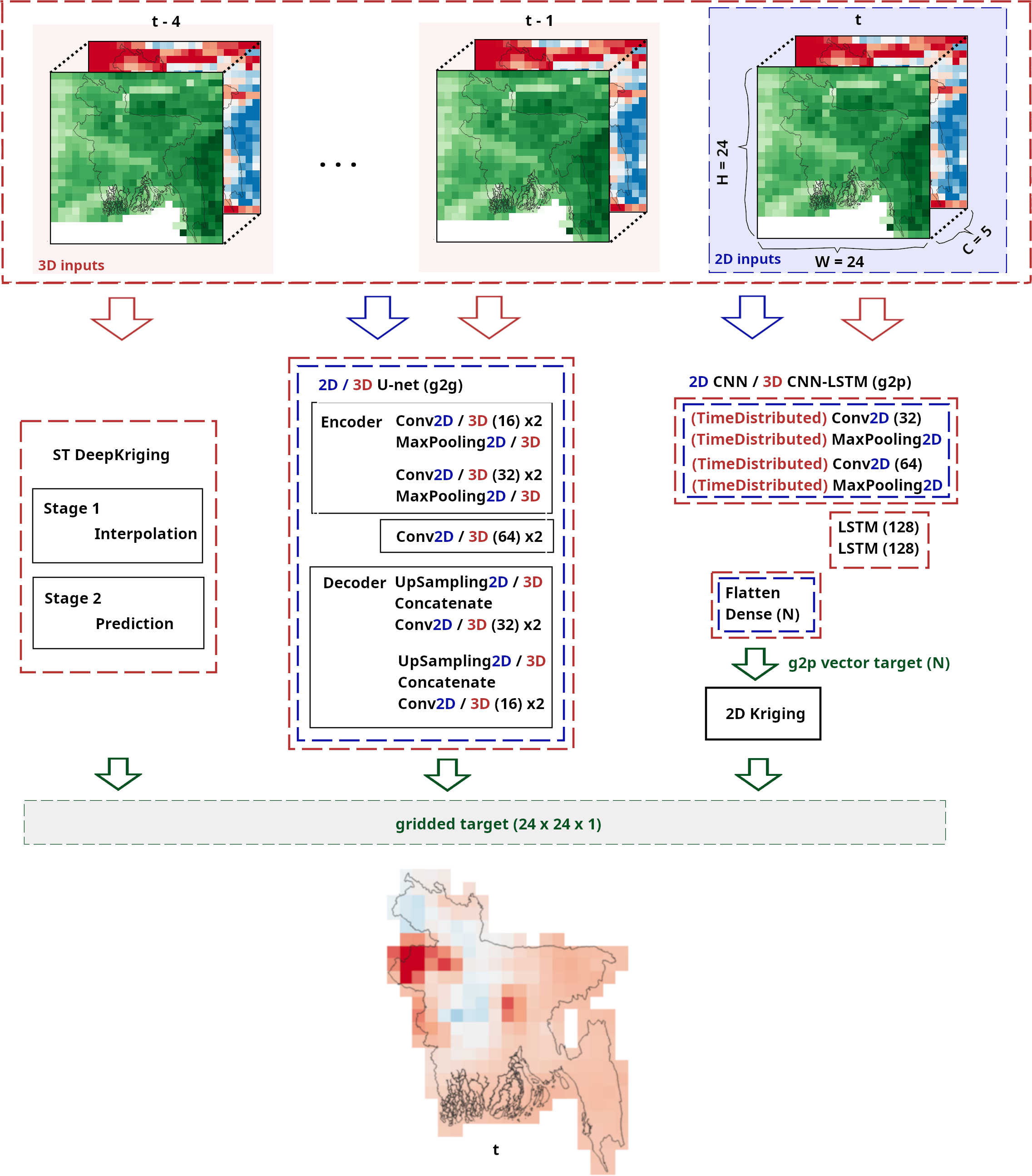}
  \caption{Employed spatial (2D) and spatiotemporal (3D) models, as described in Section \ref{methods}. 2D models are fed with gridded predictors ($C=5$) from the same timestep; 3D models additionally use lagged predictors $t...t-4$. The optimal temporal lag was found empirically during preliminary analysis. Results of all approaches are reported in Fig. \ref{fig:boxplots}. Spatiotemporal DeepKriging was implemented according to \cite{nag_spatio-temporal_2023}. Note that Stage 2 of DeepKriging only uses past predictors, as opposed to all other models.}
  \label{fig:models}
\end{figure}

Detailed model implementation is summarised in Fig \ref{fig:models}.
Table \ref{tab:performance} shows the performance of employed data-driven models, and the alignment of the GLDAS 2.2 CLSM GWS simulations with the interpolation and prediction sets. Note that GLDAS 2.2 CLSM does not follow the train-validation-test CV, and the division is only retained for comparison with data-driven models. All data-driven models clearly outperform the physical GLDAS 2.2 CLSM model in simulating changes in groundwater storage in Bangladesh, as land surface models like GLDAS are known to generalise groundwater model components, leading to diverging predictions and considerable bias \cite{gleeson_gmd_2021}. Data-driven models excel at learning spatiotemporal relationships whereas physics-driven models can generate physically consistent values and extremes beyond those seen in training data.

All models are light-weight, ranging from 118,561 (2D U-net) to 838,251 (3D CNN-LSTM stack) parameters. However, the subsequent kriging step in the latter approach adds substantial computational complexity ($O(n^2)$ space, $O(n^3)$ time). The size of our models is comparable to the combined size of 2-stage DeepKriging (\smalltilde200,000 parameters).

\begin{figure}[!ht]
  \centering
  \includegraphics[width=\textwidth]{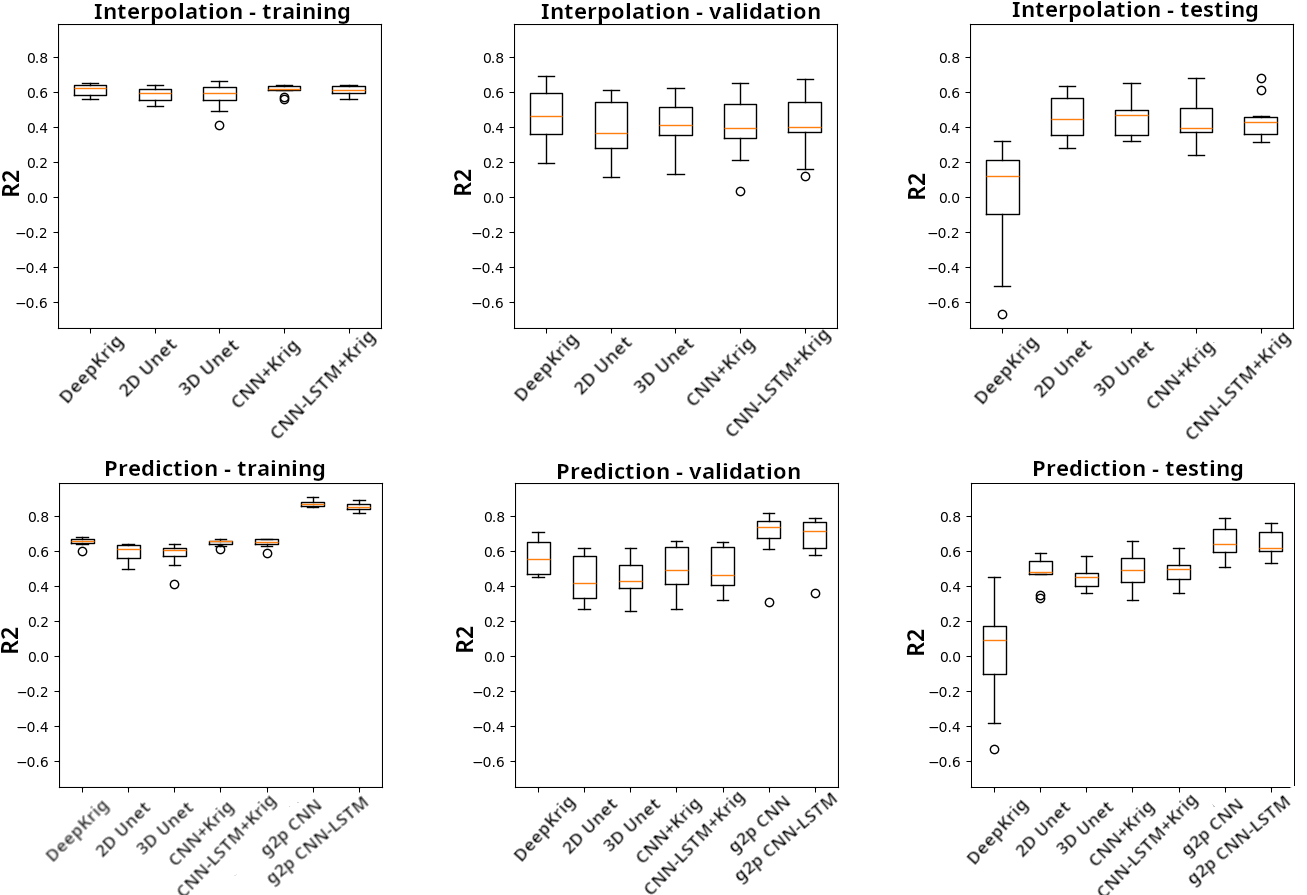}
  \caption{Performance on $\Delta$GWS dataset from Bangladesh on data-driven models, R$^2$ over 10-fold temporal CV. Refer to Tab. \ref{tab:performance} for mean performance scores.}
  \label{fig:boxplots}
\end{figure}

Poor performance of the DeepKriging model on the test sets might be explained by its simple shallow neural network architecture and its limited use of gridded predictors. The drop in performance on the test set indicates that the framework overfits training and validation data. Although other employed models rely on convolutional layers to extract spatial patterns from the gridded inputs, DeepKriging only takes into consideration scalar values. Groundwater levels often reflect changes in distant predictors, as recharge zones may lie far from where observations are made.

Upon a closer inspection of R$^2$ scores (Fig. \ref{fig:boxplots}), there is a clear distinction in performance of the individual tasks. The box plots of g2p models show a pure prediction spread, as these models don't involve interpolation. Adding kriging on top of them, and the alternative g2g approach, degrades these g2p estimates, which is marked by an increased spread in prediction in other models. At the same time, interpolation spread is consistently higher that the prediction spread of g2p models. This indicates that interpolation might be a more difficult problem, especially on the dataset that is as spatially heterogeneous as groundwater dynamics in Bangladesh. Even though the employed g2g and g2p models show minimal performance differences on the testing sets, we conclude that g2p CNN is the most suitable model for prediction, and that interpolation requires further research. This finding recognises that, as g2p CNN is one of the more sizeable models here, there is a trade-off between good performance and model size. MSE scores display a similar pattern, and can be examined in the code repository.

\begin{figure}[h]
  \centering
  \includegraphics[width=0.95\textwidth]{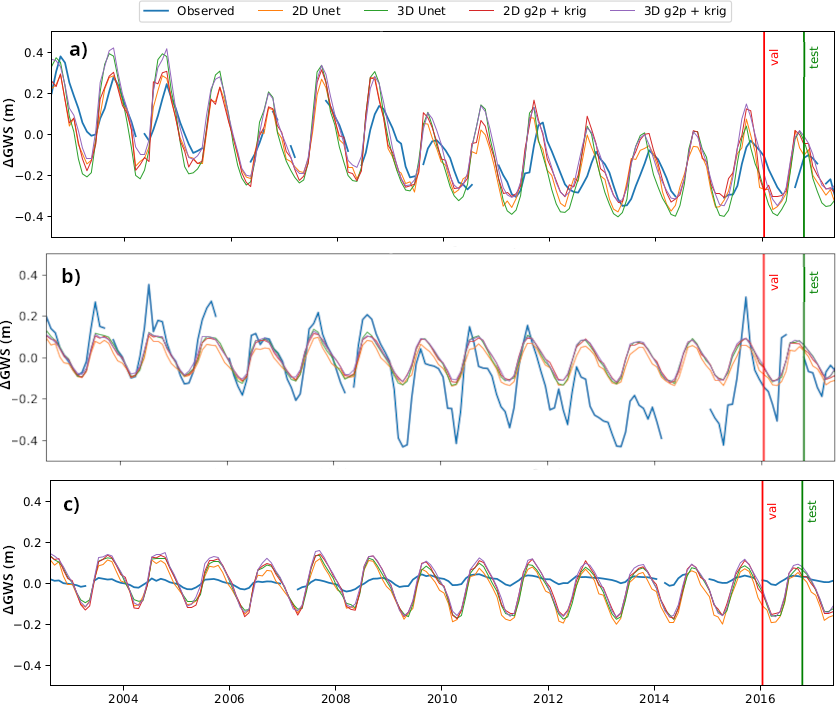}
  \caption{Performance of the employed models at three example locations: (a) RJ092 (b) RA-22 (c) DI030 from the interpolation holdout set; last CV fold (95\%-5\%-5\%).}
  \label{fig:examples_ipol}
\end{figure}

Visual inspection of the final CV fold (with a train-val-test ratio of 90\%-5\%-5\%) reveals mixed performance for the interpolation holdout dataset (points excluded prior to the 10-fold CV). As illustrated in Figure \ref{fig:examples_ipol}a, well \textit{RJ092} situated near others with comparable seasonality and trend patterns shows modelled values that align closely with observed data. For well \textit{RA-22} (Fig. \ref{fig:examples_ipol}b), the estimated values are primarily driven by its only nearby neighbour with very stable, regular seasonality. In contrast, well \textit{DI030\_1} (Fig. \ref{fig:examples_ipol}c) is surrounded by locations with stronger seasonal signals, experiencing consistent under- and overestimation of $\Delta$GWS values. Poor interpolation performance on wells \textit{RA-22} and \textit{DI030\_1} might be caused by the fact that groundwater dynamics do not always follow the first geographical principle "everything is related to everything else, but near things are more related than distant things". In complex hydrogeological settings like Bangladesh, nearby points can tap into different hydrogeological conditions. The heterogeneity of time-series dynamics, presented in Fig. \ref{fig:dist}, may occur in the neighbouring wells.

Fig. \ref{fig:boxplots} demonstrates that g2p models, mapping gridded predictors to point target, perform substantially better; interpolation of these points through kriging, as well as the alternative g2g modelling approach, degrade model fit at the observed points. Exploration of well groups reveals that the division into the regular grid cells often leads to the situation that wells that show similar patterns end up in separate grid cells, and the connection between them is lost. At the same time, wells that show very different patterns, are aggregated into the same grid cell, collectively producing a mean time series for that grid cell that is unrealistic. Therefore, image-like representation of the target variable, as required by the g2g models, may not be appropriate. The alternative g2p method that employs kriging on top of point predictions does not solve the problem described above. As kriging operates at the same grid of 0.25$^\circ$, the raw g2p (point) performance is degraded to a similar extent, regardless of whether the target averaging takes place before DL modelling (g2g method) or after (g2p with interpolation).

This finding motivates a search for an alternative interpolation technique beyond a regular grid. As subsurface geology is often unknown, with surface geology being an (imperfect) proxy, interpolation informed by clustering of locations based on their time-series dynamics, is an interesting task to explore.

Based on the current dataset and study area, there appears to be little difference in performance of the spatial (2D) vs spatiotemporal (3D) versions of the g2p and g2g models. This outcome aligns with expectations: in humid environments like Bangladesh, groundwater responds rapidly to predictors, primarily precipitation. Future research will test whether this pattern persists with independent environmental variables. 

Notably, we were unable to perform full 3D kriging on our dataset because of its prohibitive computational cost, even when employing localisation strategies. Since Gaussian processes can be viewed as a probabilistic generalisation of kriging and inherit a similarly unfavourable scaling with the number of observations, they are likewise impractical at this scale. Given this strong requirement for computational efficiency, a natural extension of this work is to explore neural processes, which operate in continuous domains while offering substantially improved scalability compared to Gaussian-process-based approaches.

\section{Conclusions}
This analysis explored the interplay of spatial interpolation and temporal prediction on an example dataset of groundwater storage changes in Bangladesh. The interpolation task was shown to be more challenging than prediction, due to the complexity of subsurface processes and lack of spatial datasets on meaningful hydrogeological variables. Employed interpolation methods - both the target point aggregation with subsequent grid inpainting using 2D/3D U-net, and point prediction with subsequent interpolation using Kriging, considerably degraded grid-to-point model fit at observed locations. Nevertheless, our models substantially outperform the estimates of groundwater storage changes by the commonly employed process-based GLDAS 2.2 CLSM model in Bangladesh. Future work is motivated by the search for an advanced interpolation technique informed by clustering of locations based on their time-series dynamics, that is generalisable to other environmental variables with complex spatial patterns.

\begin{credits}
\subsubsection{\ackname} We thank the Bangladesh Water Development Board (BWDB) for providing groundwater-level dataset.
This work was supported by a PhD studentship to AP from the London NERC Doctoral Training Programme (grant no. NE/S007229/1). RGT acknowledges support of a fellowship from the CIFAR Earth-4D program.

\end{credits}
%
%
%
\bibliographystyle{splncs04}
\bibliography{references, add_ref}

\end{document}